\documentclass[11pt,a4paper]{article}
\usepackage[hyperref]{ijcnlp2017}

\usepackage{latexsym}
\usepackage{times}
\usepackage{color}
\usepackage{amsfonts}
\usepackage{graphicx}
\usepackage{amsmath,amsfonts,amssymb,bm}
\usepackage{booktabs} 
\usepackage{array}
\usepackage{makecell}
\usepackage{paralist}
\usepackage{verbatim}
\usepackage[caption=false]{subfig}

\ijcnlpfinalcopy

\def\ijcnlppaperid{***}

\newcolumntype{L}[1]{>{\centering\let\newline\\\arraybackslash\hspace{0pt}}m{#1}}
\newcolumntype{C}[1]{>{\centering\let\newline\\\arraybackslash\hspace{0pt}}m{#1}}
\newcolumntype{R}[1]{>{\centering\let\newline\\\arraybackslash\hspace{0pt}}m{#1}}

\usepackage{boxedminipage}
\usepackage{enumitem}

\newcommand{\reduceVerticalSpace}{
\setlength{\topsep}{0pt}
\setlength{\partopsep}{0pt}
\setlength{\parskip}{0pt}
\setlength{\parsep}{0pt}
\setlength{\itemsep}{0pt}
\setlength{\itemindent}{0pt}
\setlength{\listparindent}{0pt}
}

\title{Multi-Task Learning for Speaker-Role Adaptation in\\
Neural Conversation Models}
\author{
  Yi Luan$^{\dagger*}$\quad 
  Chris Brockett$^\ddagger$ \quad 
  Bill Dolan$^\ddagger$  \quad 
  Jianfeng Gao$^\ddagger$ \quad
  Michel Galley$^\ddagger$
  \\
  $^\dagger$Department of Electrical Engineering, University of Washington\\
  $^\ddagger$Microsoft Research
  \\
  {\small {\tt luanyi@uw.edu, \{chrisbkt,billdol,jfgao,mgalley\}@microsoft.com}}
}

\date{}

\begin{document}

\maketitle

{\let\thefootnote\relax\footnotetext{* This work was performed at Microsoft.}}

\begin{abstract}
Building a persona-based conversation agent is challenging owing to the lack of large amounts of speaker-specific conversation data for model training. This paper addresses the problem by proposing a multi-task learning approach to training neural conversation models that leverages both conversation data across speakers and other types of data pertaining to the speaker and speaker roles to be modeled. Experiments show that our approach leads to significant improvements over baseline model quality, generating responses that capture more precisely speakers' traits and speaking styles. The model offers the benefits of being algorithmically simple and easy to implement, and not relying on large quantities of data representing specific individual speakers. 
\end{abstract}
\section{Introduction}

Conversational engines are key components of intelligent ``personal assistants'' such as Apple's Siri and Amazon's Alexa. 
These assistants can perform simple tasks, answer questions, provide recommendations, and even engage in chit-chats~\cite{de2008spoken,chen2015leveraging,chen2016end}. 
The emergence of these agents has been paralleled by 
burgeoning interest in training natural-sounding dialog systems from conversational exchanges between humans
\cite{ritter2011data,sordoni2015neural,luan2014relating,luan2015efficient,vinyals2015neural}. 
A major challenge for data-driven systems is how to generate output that corresponds to specific traits that the agent needs to adopt, as they tend to generate ``consensus'' responses that are often commonplace and uninteresting \cite{li2015diversity,shao2017generating}. 

\begin{figure}
\centering
\fbox{
\begin{small}
\begin{tabular}{l}
{\it \textbf{User input:} I am getting a loop back to login page.}\\
\textbf{Baseline model:} Ah, ok. Thanks for the info.\\
\textbf{Our model:}  I'm sorry to hear that. Have you tried\\
clearing your cache and cookies?\\

\end{tabular}
\end{small}
}
\caption{Existing neural conversational models (baseline) tend to produce generic responses. The system presented in this paper better represents the speaker role (support person), domain of expertise (technical), and speaking style (courteous).}
\label{fig:intro}
\end{figure}

This is illustrated in Fig.~\ref{fig:intro}, where the output of a standard Sequence-to-Sequence conversation model is contrasted with that of the best system presented in this work.  
The baseline system generates a desultory answer that offers no useful information and is  unlikely to inspire user confidence. 
The output of the second system, however, strongly reflects the agent's role in providing technical support. 
It not only evidences domain knowledge, but also manifests the professional politeness associated with a speaker in that role.

The challenge for neural conversation systems, then, is that an agent needs to exhibit identifiable role-specific characteristics (a `persona').  
In practice, however, the conversational data needed to train such systems may be scarce or unavailable in many domains. 
This may make it difficult to train a system represent a doctor or nurse, or a travel agent.
Meanwhile, appropriate non-conversational data (e.g., blog and micro-blog posts, diaries, and email) are often abundant and may contain much richer information about the characteristics of a speaker, including expressive style and the role they play. 
Yet such data is difficult to exploit directly, since, not being in conversational format, it does not mesh easily with existing source-target conversational models. 

In this paper we address the joint problems of blandness and data scarcity with multi-task learning \cite{caruana1998multitask,liu2015representation,luan2016multiplicative}.  
This is a technique that has seen success in machine translation, 
where large monolingual data sets have been used to improve translation models 
\cite{sennrich2015improving}.  
The intuition is that if two tasks are related, then joint training and parameter sharing can enable one task to benefit the other.
In our case, this sharing is between two models: On one hand, a standard Sequence-to-Sequence conversational models is trained to predict the current response given the previous context. On the other hand, using the non-conversational data, we introduce an autoencoder multi-task learning strategy that predicts the response given the same sequence, but with the target parameters tied with the general conversational model.
Our experiments with 4M conversation triples show that multi-task adaptation is effective in that the generated responses capture speaker-role characteristics more precisely than the baseline. 
Experiments  on  a  corpus  of Twitter conversations demonstrate that multi-task learning can boost performance  up  to 46.2\% in BLEU score and 23.0\% in perplexity, with a commensurate consistency gains in human evaluation.

\section{Related Work}
\subsection{Conversational Models}
In contrast with much earlier work in dialog,
our approach to conversation is wholly data-driven and end-to-end.   
In this respect, it follows a line of investigation begun by \cite{ritter2011data}, who present a statistical machine translation based conversation system. End-to-end conversation models have been explored within the framework of neural networks \cite{sordoni2015neural,vinyals2015neural,li2015diversity,li2016persona, luan2017scientific}. 
The flexibility of these Sequence-to-Sequence (\textsc{Seq2Seq}) encoder-decoder neural models opens the possibility of integrating different kinds of information beyond the single previous turn of the conversation. 
For example, \cite{sordoni2015neural} integrate additional contextual information via feed-forward neural
networks. 
\cite{li2015diversity} use Maximum Mutual Information (MMI) as the objective  function  in  order to produce more diverse and interesting responses.  
\cite{mei2016coherent} introduce an attention mechanism into an encoder-decoder network for a conversation model.

\cite{wen2015semantically} introduced a Dialog-Act component into the LSTM cell to guide generated content. 
\cite{luan2016lstm} use a multiplicative  matrix on word embeddings to bias the word  distribution of different speaker roles.
That work, however, assumes only two roles (questioner and answerer) and is less generalizable than the model proposed here.  

Most relevant to the present work, \cite{li2016persona} propose employing speaker embeddings to encode persona information and allow conversation data of similar users on social media to be shared for model training. 
That work focused on individuals, rather than classes of people. 
The approach, moreover, is crucially dependent on the availability of large-scale conversational corpora that closely match the persona being modeled\textemdash data that, as we have already observed, may not be readily available in many domains. 
In this work, we circumvent these limitations by bringing non-conversation corpora (analogous to the use of monolingual data in machine translation) to bear on a general model of conversation. 
Doing so allows us to benefit in terms of representing both the role of the agent and domain content.  

\subsection{Multi-Task Learning}
Multi-task learning has been successfully used to improve performance in various tasks, including machine translation \cite{sennrich2015improving} and image captioning \cite{luong2015multi}. 
\cite{sennrich2015improving} report methods of exploiting monolingual data---usually available in much larger quantities---to improve the performance of machine translation, including  multi-task learning of a language model for the decoder. 
Autoencoders are widely used to initialize neural networks \cite{dai2015semi}.
\cite{luong2015multi} show that an autoencoder of monolingual data can help improve the performance of bilingual machine translation in the form of multi-task learning. 
In our models, we share the decoder parameters of a \textsc{Seq2Seq} model and autoencoder to incorporate textual information through multi-task learning.

\section{Background}
\label{sec:pw}
\subsection{Task definition}
The task of response generation is to generate a response given a context. 
In this paper, following ~\cite{sordoni2015neural}, each data sample is represented as a \textit{(context,message,response)} triple, where context is the response of the previous turn, and the message is the input string of the current turn. 
The response, then, is the sequence to be predicted given these two strings of information.
In addition to the triple, large-scale  non-conversational data from the responder is provided as side information.

\subsection{Sequence-to-Sequence Conversational Models}
\label{sec:seq2seq}

Given a sequence of inputs $X=\{x_{1},x_{2},\dots,x_{n_{X}}\}$ and the corresponding output $Y=\{y_{1},y_{2},\dots,y_{n_{Y}}\}$, Sequence-to-Sequence (\textsc{Seq2Seq}) models use a Long Short-Term Memories (LSTM) \cite{hochreiter1997long} to encode the input sequence, taking the last hidden state of encoder $h_{n_X}$ to represent output sequence. The decoder is initialized by $h_{n_X}$, and predict output $y_t$ given $h_{n_X}$ and $y_{t-1}$. 

Our input is context followed by message, delimited by an \textit{EOS} token. 
The LSTM cell includes an input gate, a memory gate and an output gate, respectively denoted as $i_{t},f_{t}$ and $o_{t}$.

\subsection{Persona-based conversational model}
\label{sec:persona}
The persona-based conversational model is a variant of standard
\textsc{Seq2Seq} models, with user information encoded at decoder.
As in standard
\textsc{Seq2Seq} models, the persona-based conversational model presented in \cite{li2016persona} first encodes the source message into a vector representation using the source
LSTM. Then, for each element in the target side, hidden
units are obtained by combining the representation
produced by the target LSTM at the previous time
step $h_{t-1}$, the word representations $e_t$ at the current time
step, and the  embedding $s_i$ for user $i$. 

\begin{equation}
\label{eq:persona}
\begin{bmatrix}
  i_{t}\\
  f_{t}\\
  o_{t}\\
  l_{t}
\end{bmatrix}
=
\begin{bmatrix}
  \sigma\\
  \sigma\\
  \sigma\\
  \mathrm{tanh}
\end{bmatrix}
W \cdot 
\begin{bmatrix}
 h_{t-1}\\
 e_{t}\\
 s_i
\end{bmatrix}
\end{equation}\\[-0.5cm]
\begin{equation}
  c_{t} = f_{t}\cdot c_{t-1} + i_{t} \cdot l_{t}
\end{equation}\\[-0.5cm]
\begin{equation}
  h_{t} = o_{t} \cdot \mathrm{tanh}(c_{t})
\end{equation}

\noindent where $W \in \mathbb{R}^{4K\times 3K}$. 
This model assigns one $K$ dimensional vector representation to each of the speakers in the corpus.
It thus relies on the availability of sufficient conversational training data of each speaker to learn meaningful speaker embeddings. 
Since this type of data is usually hard to obtain in real application scenarios, we need a method that can leverage easier-to-obtain non-conversational personal data in order to incorporate richer personal information into conversational models.

\section{A Multi-task Learning Approach}

\begin{figure*}[t]
\centering
\includegraphics[width=12.5cm]{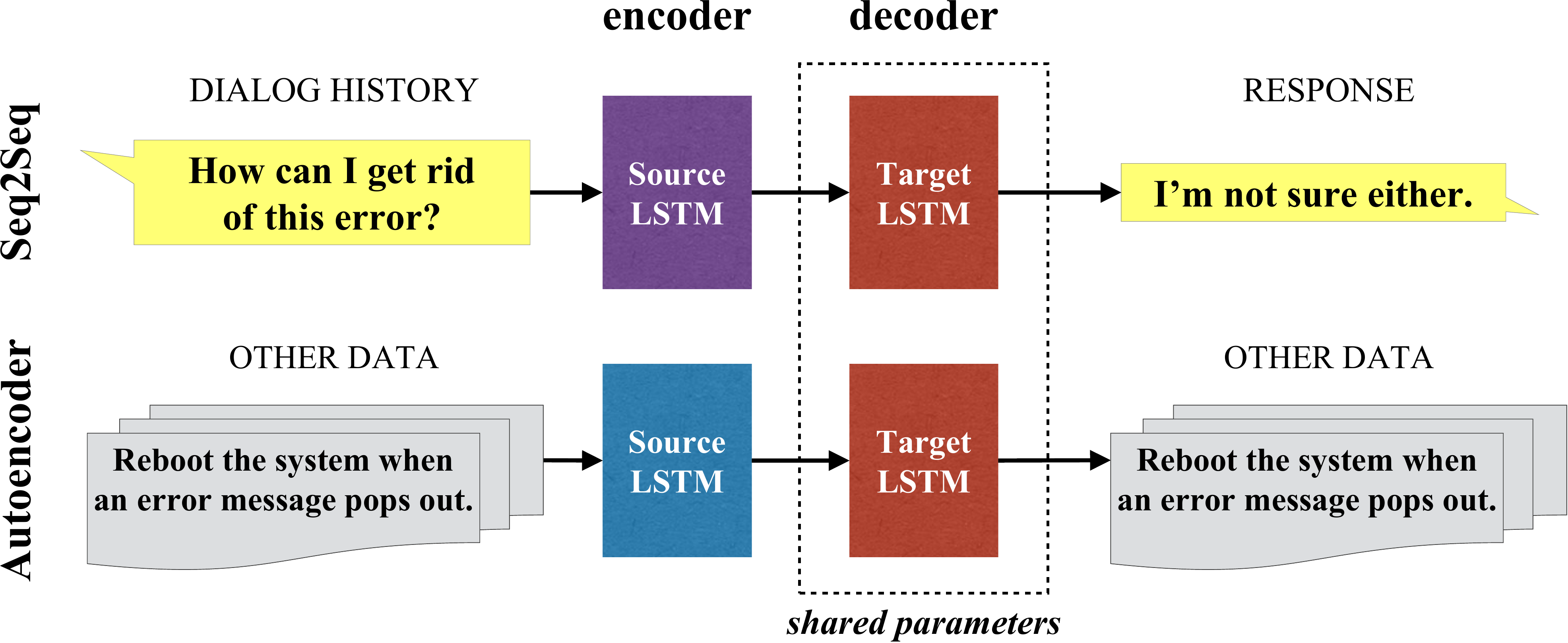}
\caption{Framework of Multi-task learning. The parameters of decoder are shared across the two tasks.}
\label{fig:multi-task}
\end{figure*}

Given the limitations of previous methods, we propose the following multi-task learning approach in order to simultaneously leverage conversational data across many users on the one hand, and personal but non-conversation data (written text) of a specific user on the other.
 We define the following two tasks:
 \begin{itemize}
\reduceVerticalSpace
\item A \textbf{\textsc{Seq2Seq}} task  that learns conversational models described in Section~\ref{sec:pw} using conversation data of a large general population of speakers. 
\item An \textbf{\textsc{Autoencoder}} task that utilizes large volumes of non-conversational personal data from target speakers.
\end{itemize}

 \paragraph{\textsc{Autoencoder}:} An \textsc{Autoencoder} is an unsupervised method of obtaining sequence embeddings based on the \textsc{Seq2Seq} framework. Like a \textsc{Seq2Seq} model, it comprises encoding and decoding components built by an LSTM sequential model as in Section~\ref{sec:seq2seq}.
Instead of mapping source to target as in a \textsc{Seq2Seq} model, the \textsc{Autoencoder} predicts the input sequence itself.

 \paragraph{Parameter sharing:} 
 Given the same context, we want to generate a  response that can mimic a particular target speaker.
Therefore, we share only the decoder parameters of \textsc{Seq2Seq} and \textsc{Autoencoder} while performing multi-task learning, so that the language model for generation can be adapted to the  target-speaker . Since the context is not constrained and can be from any speaker, the encoder parameters are not tied and are learned separately by each task. (See Fig.~\ref{fig:multi-task}.)

 \begin{figure}[t]
 \begin{minipage}[t]{0.45\textwidth}
 \fbox{
 {\small
 \parbox{\textwidth}{%
    \textbf{Training procedure of Multi-task learning:}
    \begin{compactenum}
    \item Randomly initialize \textsc{Seq2Seq} and \textsc{Autoencoder} encoder parameters.
    \item Train \textsc{Seq2Seq} model until dev set performance converges in perplexity.
    \item \textbf{While} not dev set performance converged in perplexity \textbf{do}:
    \begin{compactenum}
   \item Randomly pick a batch of samples from general conversational data.
   \item Compute loss and gradient for \textsc{Seq2Seq} task and update parameters.
    \item 
    Randomly pick a batch of samples from non-conversational data of the target user.
    \item Compute loss and gradient for \textsc{Autoencoder} task and update parameters.
    \end{compactenum}
        \item Choose the best model based on \textsc{Seq2Seq} perplexity performance on dev set.

    \end{compactenum}
  \vspace{1mm}
}
}
}
\end{minipage}

\caption{Training Procedure}

\label{fig:alg}

\end{figure}
 \paragraph{Training Procedure} The training procedure is shown in Fig.~\ref{fig:alg}. In each iteration, the gradient of each task is calculated according to the task-specific objective. The training process finishes when  perplexity performance converges in dev set and the best model is selected according to \textsc{Seq2Seq} perplexity performance.

\section{Single v.s. Multiple speaker Settings}
Two variants of \textsc{Seq2Seq} task are explored:  
\begin{itemize}
\reduceVerticalSpace
\item \textbf{\textsc{MTask-S}} Personalized response generation for a \textbf{single} user, which uses the basic \textsc{Seq2Seq} conversational model as described in Section~\ref{sec:seq2seq}. 
\item \textbf{\textsc{MTask-M}} Response generation for \textbf{multiple} users, which uses the persona-based \textsc{Seq2Seq} model described in Section~\ref{sec:persona}. 
\end{itemize}
\paragraph{\textsc{MTask-S}:} We train a personalized conversational model for one speaker at a time. For each target user, we need to perform separate multi-task training which results in $N$ models for $N$ users. This is inefficient in both memory and computational cost.
\paragraph{\textsc{MTask-M}:}In order to address the memory and computation issue of \textsc{MTask-S}, we introduce user embeddings to \textsc{Seq2Seq} model as in Eq.~\ref{eq:persona}. We first train a persona-based conversational model using conversational data for a general population of speakers. This model differs \textsc{MTask-S} in that it introduces two parameter matrices into the decoder: a speaker embedding $s_i$ and its corresponding weight matrix that can decouple speaker dependent information from general language information. In the multi-task stage, since the target users have never appeared in the  training data, we randomly initialize the user embeddings for those users and follow the training procedure as in Figure~\ref{fig:alg}.\footnote{The model can also be learned without pre-training (omitting step 2), but we found that pre-training usually helps.} The embedding of the unseen user is updated by \textsc{Autoencoder} training together with the decoder LSTM parameters.

\section{Experimental Setup}

\subsection{Datasets}

As training data, we use a collection of 3-turn conversations  extracted from the Twitter FireHose. 
The dataset covers the six-month period beginning January 1, 2012, and was limited to conversations where the responders had engaged in at least 60 3-turn conversational Twitter interactions during the period. In other words, these are people who reasonably frequently engaged in conversation, and might be 
experienced ``conversationalists.''

We selected the top 7k Twitter users who had most conversational data from that period (at least 480 turns, average: 571). 
This yielded a total of approximately 4M conversational interactions. 
In addition to these 7k general Twitter users, we also selected the 20 most frequent users, employing all of their conversation data for development and test. Twitter users typically have many more single posts than posts that interact with other people. 
We therefore treat single posts as non-conversation data. All single posts of the 20 top users (at least 9k per user, average 10.3k) were extracted for multi-task learning.  
The 20 users were of diverse backgrounds, including technical support personnel, novelists, and sports fans.

\subsection{Evaluation}

As in previous work \cite{sordoni2015neural},
we use BLEU and human evaluation for evaluation. 
BLEU \cite{papineni2002bleu} has been shown
to correlate fairly well with human judgment at a document- and corpus-level, including on the response generation task.\footnote{\cite{liu:2016} suggest that BLEU doesn't correlate well with human judgment at the sentence level. Other work, however, has shown 
that correlation increases substantially with larger units of analysis (e.g., document or corpus) \cite{galley2015,MetricsMATR:2008}.}
We also report perplexity as an indicator of model capability. 

We additionally report degree of diversity by calculating
the number of distinct unigrams and bigrams
in generated responses. The value is scaled by total
number of generated tokens to avoid favoring long
sentences (shown as distinct-1 and distinct-2). Finally, we present a human evaluation that validates our 
main findings.

\subsection{Baseline}

Our baseline is our implementation of the
LSTM-MMI of \cite{li2015diversity}. 
The MMI algorithm reduced the blandness of \textsc{Seq2Seq} models
by scoring the generated N-best list with a function that linearly combines a length
penalty and the log likelihood 
of source given target:
\begin{equation}
\log p(R|M, v) + \lambda \log p(M|R) + \gamma|R| 
\end{equation}
where $p(R|M, v)$ is the probability of the
generated response given message $M$ and the
respondent’s user 
ID. $|R|$ is the length
of the target and $\gamma$ is the associated penalty
weight. We use MERT 
\cite{och2003minimum}
to optimize $\gamma$ and $\lambda$ on BLEU using N-best lists of
response candidates generated from the development
set. 
To compute $p(M|R)$, we train an inverse
\textsc{Seq2Seq} model by swapping messages and responses.
The reverse \textsc{Seq2Seq} models 
$p(M|R)$ is trained with no user information considered. 

\subsection{Training and Decoding}

We trained two-layer \textsc{Seq2Seq} models on the Twitter corpus, using the following settings:
\begin{itemize}
\reduceVerticalSpace
\item 2 layer LSTM models with 500 hidden cells for each layer.
\item Batch size is set to 128.
\item Optimization method is Adam~\cite{kingma2014adam}.
\item Parameters for \textsc{Seq2Seq} models are initialized by sampling from uniform distribution $[-0.1,0.1]$.
\item Vocabulary size is limited to 50k.
\item Parameters are tuned based on perplexity.
\end{itemize}

For decoding, the N-best lists are generated with beam size $B=50$. 
The maximum length of the generated candidates was set at 20 tokens. 
At each time step, we first examine all $B\times B$ possible next-word candidates, and add all hypotheses ending with an $EOS$ token to the N-best list. We then preserve the top-B unfinished hypotheses and move to the next word position. We then use LSTM-MMI to rerank the N-best list and use the 1-best result of the re-ranked list in all evaluation.

\begin{table}
  \centering
 {\small
   \begin{tabular}{rrrr}
    \toprule

     & Baseline & \textsc{MTask-S} & \textsc{MTask-M} \\
     \midrule
            Perplexity & 56.33 & {\bf 32.27} & 44.96\\
           (dev) & & {\small (-42.7\%)} & {\small (-20.2\%)}\\
            Perplexity & 61.17 & {\bf 39.83}             & 43.21\\
           (test) & & {\small (-34.9\%)} & {\small (-29.4\%)}\\
    \bottomrule
  \end{tabular}}
  \caption{Perplexity for standard \textsc{Seq2Seq} and the user
model on the Twitter Persona dev set.}
  \label{tab:perplexity}
\end{table}

\begin{table}
  \centering
 {\small
   \begin{tabular}{rrrr}
    \toprule
            & Baseline & \textsc{MTask-S} & \textsc{MTask-M} \\
     \midrule
     BLEU          & 1.32      &            1.76    & {\bf 2.52}\\
     (dev)         &           & {\small (+33.3\%)} & {\small (+90.1\%)}\\
     BLEU          & 1.31      &            1.69    & {\bf 2.25}\\
     (test)        &           & {\small (+29.0\%)} & {\small (+71.7\%)}\\
     \midrule
     distinct-1    & 1.69\%    &       2.43\%       & {\bf 2.44}\%\\
     distinct-2    & 6.53\%    & {\bf 10.2}\%       &      9.79\%\\
    \bottomrule
  \end{tabular}}
  \caption{Performance on the Twitter dataset of 2-layer \textsc{Seq2Seq} models and MMI models. Distinct-1 and distinct-2
are respectively the number of distinct unigrams and bigrams divided by total number of generated words.}
  \label{tab:BLEU}
\end{table}

\begin{figure*}%
\centering
  \includegraphics[width=13.5cm]{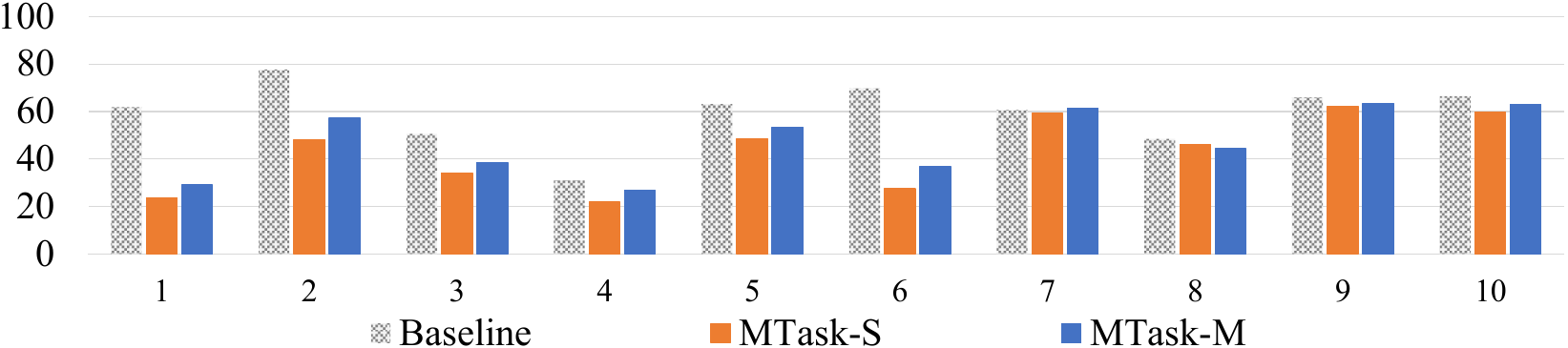}%
\caption{Perplexity scores for the top 10 users with most (non-conversational) training data. 
Users with obvious speaking styles or stronger user role characteristics (e.g., 1, 2, and 6) show much greater perplexity reduction than the other ones.}
\label{fig:results}
\end{figure*}

\section{Experimental Results}

The perplexity and BLEU score results for three models are shown in Tables \ref{tab:perplexity} and \ref{tab:BLEU}. Compared with the baseline model LSTM-MMI, we obtain a 34.9\% decrease in perplexity for the \textsc{MTask-S} model and a 29.4\% decrease in perplexity for the \textsc{MTask-M} model. Significant gains are obtained in BLEU score as well: \textsc{MTask-S} gains 29.0\% relative increase compared with the baseline and \textsc{MTask-M} gains 71.7\%. \textsc{MTask-S} performance is better than \textsc{MTask-M} in perplexity, but worse on BLEU score. Apart from the fact that BLEU does not necessarily correlate with perplexity, this result also indicates that \textsc{MTask-S} has more parameters (each user has a unique model for \textsc{MTask-S}) but tends to overfit on development set perplexity. Another possible reason that \textsc{MTask-M} performs better than \textsc{MTask-S} is the introduction of user embeddings. The persona-based conversational model can decouple the personalized information from general language patterns and can therefore encode user characteristic better.  
We further report degree of diversity by calculating the number of distinct unigrams (distinct-1) and bigrams (distinct-2) in generated responses as in Table \ref{tab:BLEU}. 
To avoid biasing toward longer sentences, this value is scaled by the total number of tokens generated.
Both \textsc{MTask-S} and \textsc{MTask-M} models perform better than baseline in terms of distinct-1 and distinct-2, which we interpret to mean that our approach can help the system generate responses that are more diverse yet better approximate the targeted speaker or speaker type. 

Fig.~\ref{fig:results} shows the perplexities for the 10 individual users most represented in the non-conversational training data. Our multi-task approaches consistently outperform baseline on perplexity.
However, the performance between individual target users can vary substantially.\footnote{We do not report BLEU scores for individual users, as the dev and test set for each specific user tends to be small (less than 500 samples) and BLEU is known to be unreliable when evaluated on small datasets \cite{graham:2015,liu:2016}.}
 
After inspecting dev set outputs, we observe that users with obvious speaking styles or stronger user role characteristics show much greater gain than the others. For example, User 1 is a technical support worker who answers web questions for Twitter users, while User 2 always expresses strong feelings and uses exclamation marks frequently. Conversely, tweets from users that did not show significant gain appear to be more about daily life and chitchat, with no strong role characteristics (e.g., Users 3 and 4). We present example outputs for User 1 and 2 in Section~\ref{sec:Discussion}. 

\subsection{Human Evaluation}

Human evaluation of the outputs was performed using crowdsourcing.\footnote{Two outputs were removed from the datasets owing to offensive content in the examples.}
Evaluation took the form of a preference test in which judges were presented with a random sample of 5 tweets written by the targeted user as example texts, and asked which system output appeared most likely to have been produced by the same person.
A 5-point scale that permitted ties was used, and system pairs were presented in random order. 
A short input message (the input that was used to generate the outputs) was also provided. 
We used 7 judges for each comparison; those judges whose variances differed by more than two standard deviations from the mean variance were discarded.  
Table \ref{tab:UHRS} shows the results of pairwise evaluation, along with 95\% confidence intervals of the means.  
\textsc{MTask-S} and \textsc{MTask-M} both perform better on average than LSTM-MMI, consistent with the BLEU results. 
\textsc{MTask-M}'s gain over the LSTM-MMI baseline is significant at the level of $\alpha = 0.05$ (p = 0.026), indicating that judges were better able to associate the output of that model with the target author.

\begin{table}
  \centering
 {%
   \begin{tabular}{lrr}
    \toprule
        & Baseline & System\\
     \midrule
 
\textsc{MTask-S} &  0.491 $\pm$0.011 &  0.504 $\pm$0.011\\
\textsc{MTask-M}  &  0.486 $\pm$0.012 &  0.514 $\pm$0.012\\

    \bottomrule
  \end{tabular}}
  \caption{Results of human evaluation, showing relative gain of \textsc{MTask-S} and \textsc{MTask-M} systems over the LSTM-MMI baseline in pairwise comparison, together with 95\% confidence intervals of the means.}
  \label{tab:UHRS}
\end{table}

\begin{figure}
\centering 
\includegraphics[width=0.47\textwidth]{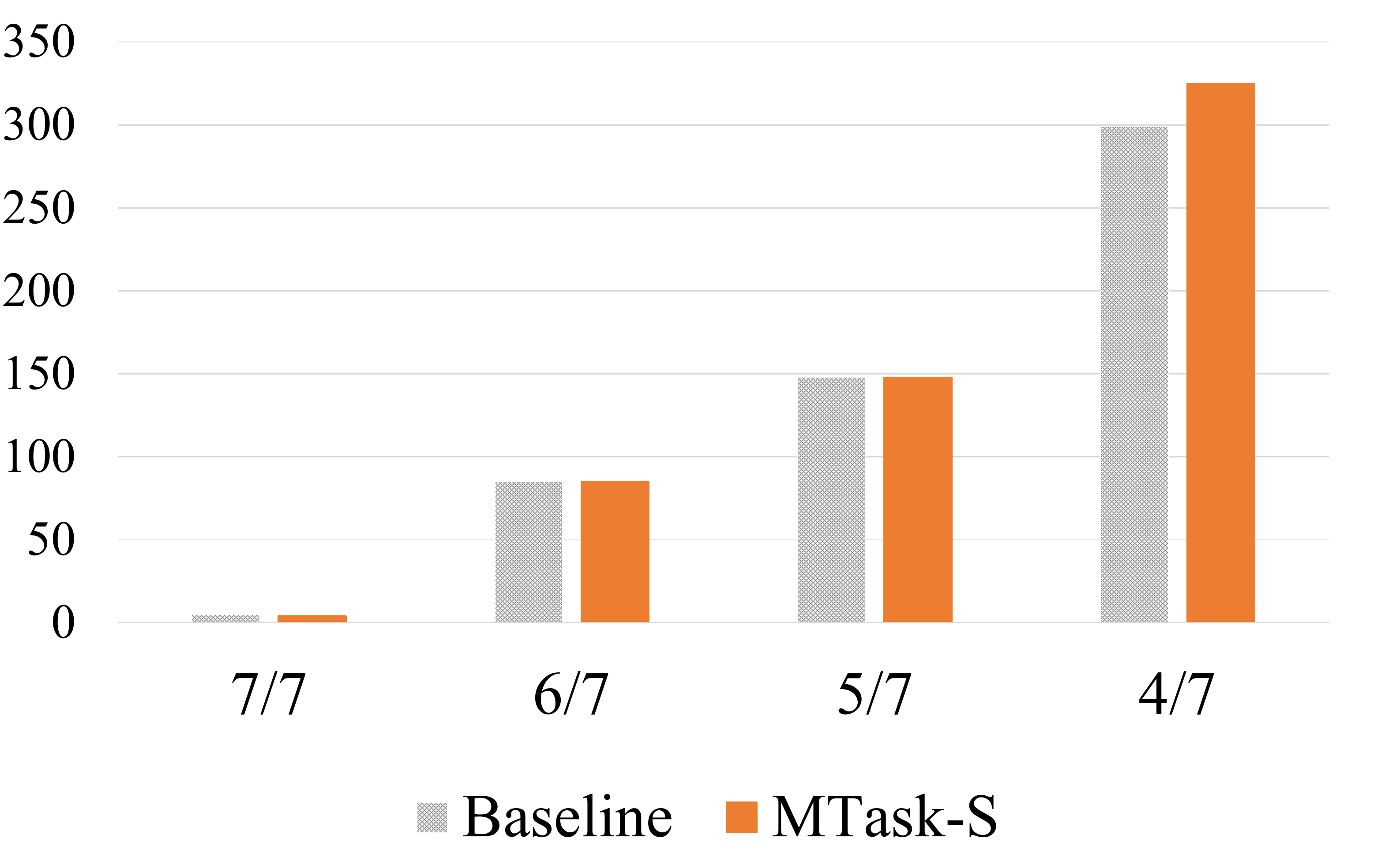}
      \caption{Judge agreement counts for \textsc{MTask-S} versus Baseline.
      The difference between the two systems is statistically significant.
      } 
      \label{fig:single-bins}
\vspace{0.15in}

\includegraphics[width=0.47\textwidth]{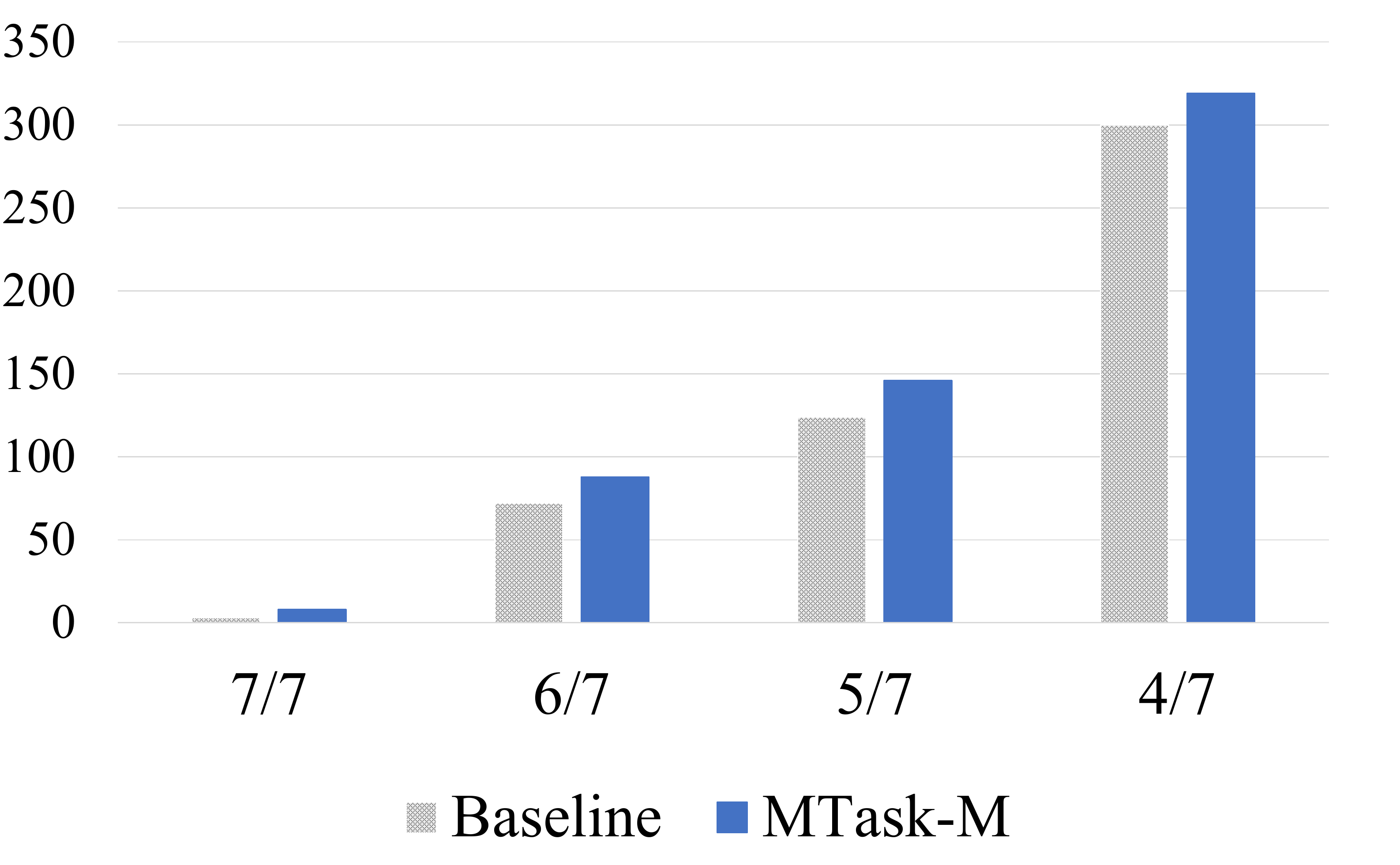}
      \caption{Judge agreement counts for \textsc{MTask-M} versus Baseline. 
      The difference between the two systems is statistically significant,
      and the preference for \textsc{MTask-M} is more evident.
      } 
      \label{fig:multi-bins}
 \vspace{-0.15in}
 \end{figure}

In Table \ref{tab:UHRS} the strength of the trends is obscured by averaging. 
We therefore converted the scores for each output into the ratio of judges who selected that system for each output (Figs. \ref{fig:single-bins} and \ref{fig:multi-bins}).
To read the charts, bin 7 on the left represents the case where all 7 judges ``voted'' for the system, bin 6 the case where 6 out of 7 judges ``voted'' for the system, and so forth.\footnote{Partial scores were rounded up. This affects both systems equally.} 
Bins 3 through 0 are not shown since these are a mirror image of bins 7 through 4. 
It can be seen that judge support for \textsc{MTask-M} (Figure \ref{fig:multi-bins}) tends to be stronger than for \textsc{MTask-S} (Figure \ref{fig:single-bins}). 

These differences are statistically significant, but they also suggest that this was a challenging task for crowd workers. In many cases, the 5 random examples may not have been sufficed to distinguish individual styles,\footnote{We limited the number to 5 with the intention of not overwhelming judges with too much information, which may have exacerbated the difficulty.} and even when distinctive, similar outputs from arbitrary inputs may not be undesirable\textemdash indeed, different individuals may legitimately respond similarly to the same input, particularly when the input itself is bland or commonplace. %

\section{Discussion}
\label{sec:Discussion}

\begin{figure*}[ht]
\centering
\begin{small}
\begin{tabular}{l}\toprule
{\it Context: have you tried uninstalling nike+ connec? how about docking it to a different computer?}\\
{\it Input: Tried both of those techniques. But nothing works. I can try the other computer again.}\\
{\bf Baseline:} Let me know if there's anything else I can help you with. Thanks for the help.\\
{\bf Multitask:} Have you tried using a different browser? Let us know if you have any other questions.\\
\midrule
{\it Context: What do you see when you have a failed login attempt? Never ending spinner? Loop you back to login page?}\\
{\it Input: I am getting a loop back to login page, windows.}\\
{\bf Baseline:} Ah, ok. Thanks for the info. I'll check it out. Thanks for the feedback.\\
{\bf Multitask:} I'm sorry to hear that. Have you tried clearing your cache and cookies?\\
\midrule
{\it Context: To clarify: your first run after resetting your connection will not post to Facebook. Does this help? Let us know.}\\
{\it Input: That's correct. I reset it twice! It still doesn't work.}\\
{\bf Baseline:} Let me know if there's anything else I can help you with. Thanks for the update.\\
{\bf Multitask:} I'm sorry to hear that. Are you receiving any error messages? Let us know.\\
\midrule
{\it Context: To link your sensor to your iPod, from home screen go to settings$>$Nike+ iPod$>$sensor$>$link new. Any luck?}\\ %
{\it Input: not work.. Any idea?}\\
{\bf Baseline:} I'm not sure, I'll check it out when i get home. Thanks for the update\\
{\bf Multitask:} Let me know if you have any other questions. Let us know if the issue persists.\\
\midrule
{\it Context: You're welcome. What generation of the iPod Nano are you using?}\\
{\it Input: 5 years old and bought the kit then but my new Nano touch also not detecting sensor maybe receiver broken}\\
{\bf Baseline:} Ah , I see. I'll have to check it out. Thanks for the update.\\
{\bf Multitask:} Ah, I see. I see. Have you been able to try a different network?\\
\midrule
{\it Context: Unfortunately you can't transfer your Nike+ training program to your iPod at this time. Let us know if this helps!}\\
{\it Input: Oh! That's a pity I bet I'd be able to run a couple more if this nice voice, told me to speed up in-between}\\
{\bf Baseline:} That's what I'm hoping for, but I don't know what to do with it.\\
{\bf Multitask:} I'm sure you will be able to run a marathon at least once a week or so.\\
\midrule\\[-0.3cm]
\midrule
{\it Context: So when are you going to post more n/d stuff?}\\
{\it Input: Soon I guess I have a chapter written, but it's way too short and not good enough to post just yet.}\\
{\bf Baseline:} Let me know how you get on with it. I don't know what to write.\\
{\bf Multitask:} Ah, I see. Well, I'm sure it'll be interesting to see what happens!\\
\midrule
{\it Context: Ah, no. Had kind of a crazy day, and I've been writing a lot tonight. I have a new one-shot almost finished!}\\
{\it Input: Alright. I'm sure you'll read it eventually}\\
{\bf Baseline:} I don't think I'll be able to sleep tonight, but I'm going to sleep.\\
{\bf Multitask:} I hope so! I can't wait to read it, hehe.\\
\bottomrule
\end{tabular}
\end{small}
\caption{Responses produced by our baseline and multi-task neural response generation system (Multi-S), given a conversation history (context and input message.) The first 6 examples and last 2 examples correspond respectively to Users 1 (tech support) and 2.}
\label{fig:examples}
\end{figure*}

Fig.~\ref{fig:examples} presents responses generated by baseline and multi-task (\textsc{MTask-M}) response generation systems. 
Both systems are presented with a conversation history of up to two dialog turns (context and input message), and this larger context helps produce responses that are more in line with the conversation flow \cite{sordoni2015neural}. The first six  response examples 
are generated for the same underlying speaker (a technical support person, User 1 in Fig.~\ref{fig:results}). The two last multi-task responses are generated for User 2.

We notice striking differences between the baseline and the multi-task model.
The six first responses of \textbf{Multitask} in Fig.~\ref{fig:examples} represent a very consistent register in three different aspects. First, it is relatively clear from these responses that the underlying speaker represented by the model is a tech support employee.
Interestingly, this employee appears to give help with fitness-related software, and responds that the customer will be ``{\it able to run a marathon}''.
On the other hand, the output of the baseline system is relatively bland and deflective. More crucially the baseline does a relatively poor job producing content words that are relevant to the speaker's domain of expertise. 
Finally, the tone of the baseline system is often incongruous, e.g., when it responds ``{\it I don't know what to do with it}'', which is unlikely to be a desirable response to offer a customer. In another case, the baseline responds ``{\it thanks for the info}''. While this kind of response is appropriate for many speaker roles, it is less appropriate here, as the support employee is the one to give information and helpful advice.

The figure also 
illustrates current limitations
of our speaker role model. For example, our response in the fourth example shows that such systems can be deflective (e.g., not giving any suggestion in response to ``{\it any idea}''), but at least the system does respond in a customer-support register.  In the fifth example, response of the system seems relatively irrelevant, but this kind of natural language comprehension problem seems almost unavoidable. Semantic congruity aside, the response strikes the right tone---it is pragmatically and socially appropriate, which is the primary purpose of this investigation. 
The final two examples of Fig.~\ref{fig:examples} show that the model is also able to learn a voice or register for a completely different kind of character. The underlying person is highly assertive---reflected in their use of exclamation marks---and speaks informally (e.g., ``hehe''), in a way the tech support person would typically not.

\section{Conclusion}
This paper introduces a multi-task learning approach to incorporate speaker role characteristics into conversational models using non-conversational data. 
Both models presented here are relatively simple to implement, and show significant improvement in perplexity and BLEU score over a baseline system. 
Overall, \textsc{MTask-M} is more computationally efficient, and effective in generating speaker-role-specific information, as reflected in human evaluation.
Responses generated by these models exhibit a marked ability to capture speaker roles, expressive styles and domain expertise characteristic of the targeted user, without heavy recourse to an individual speaker's conversational data. %

\section*{Acknowledgements}
We thank  Marjan Ghazvininejad, John Wieting,  Vighnesh Shiv,  Mari Ostendorf and Hannaneh Hajishirzi for helpful suggestions and discussions.

\raggedbottom

\bibliography{references}
\bibliographystyle{ijcnlp2017}

\end{document}